\providecommand{\paperoptions}{ustc,normalcite}
\documentclass[\paperoptions]{meta}
\microtypesetup{expansion=false}

\usepackage{xspace}
\usepackage{amsmath,amssymb}
\usepackage{tabularx,array,colortbl}
\usepackage{float}
\usepackage{wrapfig}
\usepackage{xurl}
\usepackage[most]{tcolorbox}

\newtcolorbox{suppbox}[1]{
  enhanced,
  breakable,
  colback=black!3,
  colframe=black!55,
  colbacktitle=black!72,
  coltitle=white,
  title={#1},
  fonttitle=\bfseries\small,
  boxrule=0.45pt,
  arc=1.2mm,
  outer arc=1.2mm,
  left=1.5mm,
  right=1.5mm,
  top=1.1mm,
  bottom=1.1mm,
  before skip=5pt,
  after skip=6pt
}

\title{BM25 Wins at Scale: A Scaling Study of Retrieval-Augmented Generation Paradigms}
\author[1,\ast]{Pengyu Wang}
\author[1,2,\dagger]{Benfeng Xu}
\author[1]{Shaohan Wang}
\author[1]{Mingxuan Du}
\author[3]{Xin Zeng}
\author[3]{Huarui Wu}
\author[1]{Lei Zhang}
\author[1]{Licheng Zhang}
\affiliation[1]{University of Science and Technology of China, Hefei, China}
\affiliation[2]{Metastone Technology, Beijing, China}
\affiliation[3]{Information Technology Research Center, Beijing Academy of Agriculture and Forestry Sciences, Beijing, China}
\contribution[\dagger]{Project Lead}
\abstract{Retrieval-augmented generation (RAG) spans lexical and dense retrieval,
graph-based indexing, and agentic search, but these paradigms are usually
evaluated on different benchmarks at one corpus size, leaving their
accuracy--cost scaling unclear. To bridge this gap, we present a controlled
study that varies corpus size along 28 strictly nested tiers spanning roughly
450-fold, while holding questions and a fixed bedrock of relevant and
adversarial documents unchanged. Under one reader model and one judging
protocol, we measure official accuracy, construction and query tokens, and
latency. The results reveal a scale-dependent crossover rather than an
unconditional winner. File-System Agent leads at the smallest shared tiers,
but its sequential exploration costs 39 times more query tokens at the
bedrock and becomes less effective as the search space grows. Around
10 million corpus tokens, BM25 overtakes it and leads at every larger shared
tier, with a margin approaching 20 points at full scale. BM25 also anchors
the low-cost end of the Pareto frontier without LLM-based construction.
Dense retrieval remains efficient but less accurate, whereas graph-based
RAG encounters construction walls before deployment scale and its scalable
variants remain below BM25 at shared tiers. Overall, corpus growth
increasingly favors global candidate ranking: lexical retrieval is the
strongest scalable default, while agentic reasoning works best after ranked
discovery rather than in place of it.
}
\date{July 30, 2026}
\correspondence{Licheng Zhang at \email{zlczlc@mail.ustc.edu.cn}}

\begin{document}
\maketitle
\begingroup
\renewcommand{\thefootnote}{\fnsymbol{footnote}}
\footnotetext[1]{Work done during the internship at Metastone.}
\endgroup

\section{Introduction}
Retrieval-augmented generation grounds the outputs of large language
models in external corpora, mitigating
hallucination~\citep{lewis2020retrieval}. Its methods have diverged into
paradigms whose costs arise at different stages and in different
forms. Lexical retrieval~\citep{robertson2009probabilistic} and dense
retrieval~\citep{karpukhin2020dense} require little preparation: indexing
takes at most an embedding pass over the corpus. Graph-based RAG,
including MS-GraphRAG~\citep{edge2024local},
LightRAG~\citep{guo2024lightrag}, and
HippoRAG~2~\citep{gutierrez2024hipporag,gutierrez2025rag}, invests heavily at
indexing time: it runs an LLM over every chunk to extract entities and
relations, so that the resulting structure can be exploited at query time;
LinearRAG~\citep{zhuang2025linearraglineargraphretrieval} builds the same
kind of graph with a lightweight named-entity recognizer and embeddings
instead.

\noindent
\begin{minipage}[t]{0.455\textwidth}
\vspace{0pt}
File-System Agent spends its cost at query time, using an LLM to
search the corpus through iterative file-system tool
calls~\citep{yao2023reactsynergizingreasoningacting}. Each call is
conditioned on earlier results, making the tool loop a sequential retrieval
policy. The same file-and-command interface is
established in coding-agent
systems~\citep{jimenez2024swe,yang2024sweagentagentcomputerinterfacesenable,wang2025openhandsopenplatformai}
and is used by leading industrial agents such as Claude Code and Codex.
We instantiate this interface over the corpus' raw per-source file tree,
requiring no retrieval index. Agents operating over graph indexes are
covered by our access-layer experiments.

However, far less is known about how these paradigms scale. Each
is typically evaluated on its own benchmark at a single corpus size,
whereas deployed corpora such as enterprise knowledge bases hold
hundreds of thousands of documents and keep growing. By scaling we mean
how a paradigm behaves as the corpus grows while the workload stays
fixed.
\end{minipage}\hfill
\begin{minipage}[t]{0.515\textwidth}
\vspace{0pt}
\centering
\includegraphics[width=\linewidth]{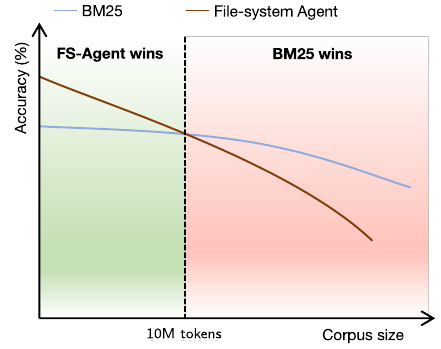}
\captionsetup{width=\linewidth,hypcap=false}
\captionof{figure}{Schematic of the observed scaling crossover. The point
estimates cross at roughly 10M corpus tokens, after which BM25 leads at
every larger shared tier. Figure~\ref{fig:acc} reports the measured curves
and confidence bands.}
\label{fig:teaser}
\end{minipage}
\par\medskip

Question-relevant and adversarial evidence is also fixed from the
smallest tier onward. We measure it on three axes: answer accuracy,
offline cost in construction tokens (generative and embedding), and
online cost in query tokens and latency. Measuring it accurately is
hard, because accuracy differences are easily confounded
by the reader model, the judge, the question set, and corpus difficulty.
Existing comparisons typically fix the corpus at a single size and vary
these factors freely, so the scaling question has remained open.

To bridge this gap, we conduct a controlled corpus-scaling study on
EnterpriseRAG-Bench~\citep{sun2026enterpriseragbenchragbenchmarkcompany},
an enterprise corpus of 511,959 documents with 500 questions and
per-question adversarial distractors. Scale is varied along a ladder of
28 strictly nested tiers, growing 1.25 times per rung from 1,144 to
511,959 documents (1.7M to 601M tokens), while question-relevant documents
and distractors remain fixed. Additional documents follow one fixed,
source-and-noise-stratified order. Every paradigm answers the same questions
using the same underlying reader model, with token-level cost metering, and
all predictions are scored under the benchmark's official protocol, with
robustness checked using an independent judge and a binary protocol. The
central finding, summarized in Figure~\ref{fig:teaser}, is a scale-dependent
crossover rather than an unconditional win at every corpus size. The
File-System Agent has the higher official point estimates at the smallest
measured shared tiers. Around 10M tokens, BM25 catches it; BM25 then leads
at every larger shared tier, while its query cost remains nearly
scale-invariant and the File-System Agent pays increasingly for sequential
exploration. At the 601M-token full corpus, the gap reaches nearly 20
points. Dense retrieval remains below both, while graph-based RAG either
reaches an early construction ceiling or remains below BM25 at the shared
tiers it completes.

Our contributions are as follows.
\begin{itemize}
\item A reusable 28-tier, 450-fold corpus ladder from 1,144 to 511,959
documents that holds questions, evidence, and distractors fixed.
\item A unified evaluation of seven pipelines across four RAG paradigms,
with token-level cost metering and matched controls.
\item A scale-dependent crossover: File-System Agent leads early, while
BM25 overtakes it around 10M corpus tokens.
\end{itemize}

\section{Related Work}
\subsection{Retrieval-Augmented Generation}
RAG grounds generation in
retrieved text through a learned or frozen reader
\citep{lewis2020retrieval,guu2020retrieval,ram2023context}; adaptive
variants further decide when to retrieve or critique evidence
\citep{asai2024self,jiang2023active}. Retrieval ranges from lexical
BM25~\citep{robertson2009probabilistic,lin2021pyserini} to learned
sparse, dense, and late-interaction models
\citep{formal2021splade,karpukhin2020dense,khattab2020colbert}. We use
BM25 and compact chunk embeddings as representative lexical and dense
interfaces, while holding the reader fixed.

\subsection{Graph-Based RAG}
These methods construct explicit structure
before answering. MS-GraphRAG builds hierarchical entity communities and
reports~\citep{edge2024local}; LightRAG indexes entities and relations
\citep{guo2024lightrag}; and HippoRAG~2 retrieves through query-linked
facts and Personalized PageRank
\citep{gutierrez2024hipporag,gutierrez2025rag}. LinearRAG instead builds
an entity co-occurrence graph with lightweight NER and embeddings, without
generative build calls
\citep{zhuang2025linearraglineargraphretrieval}. This family also includes
tree-, curated-KG-, and GNN-based indexes
\citep{sarthi2024raptorrecursiveabstractiveprocessing,chen2024plan,mavromatis2024gnn,peng2025graph},
but published evaluations generally remain at a fixed, relatively small
corpus size.

\subsection{Agentic Retrieval}
A third direction replaces the one-shot
pipeline with an LLM agent that interleaves reasoning and tool
calls~\citep{yao2023reactsynergizingreasoningacting,schick2023toolformer,qin2024toolllm}, searches
the open web~\citep{nakano2021webgpt}, retrieves adaptively during
multi-step reasoning~\citep{trivedi2023interleavingretrievalchainofthoughtreasoning,press2023measuring}, reflects
on failures~\citep{shinn2023reflexion}, or learns a search policy with
reinforcement learning~\citep{jin2025search}. These systems vary both the
controller and the retrieval substrate, whereas our scaling question
requires the controller to remain fixed. We therefore instantiate the
paradigm in the form established by coding agents: an identical tool loop
navigating a corpus through plain file-system operations
\citep{jimenez2024swe,yang2024sweagentagentcomputerinterfacesenable,wang2025openhandsopenplatformai}.
Our File-System Agent applies this interface to the raw enterprise corpus;
the substrate experiment then holds its policy model, prompt, tool loop,
and budget fixed while replacing raw files with graph indexes.

\begin{figure}[t]
\centering
\includegraphics[width=\textwidth]{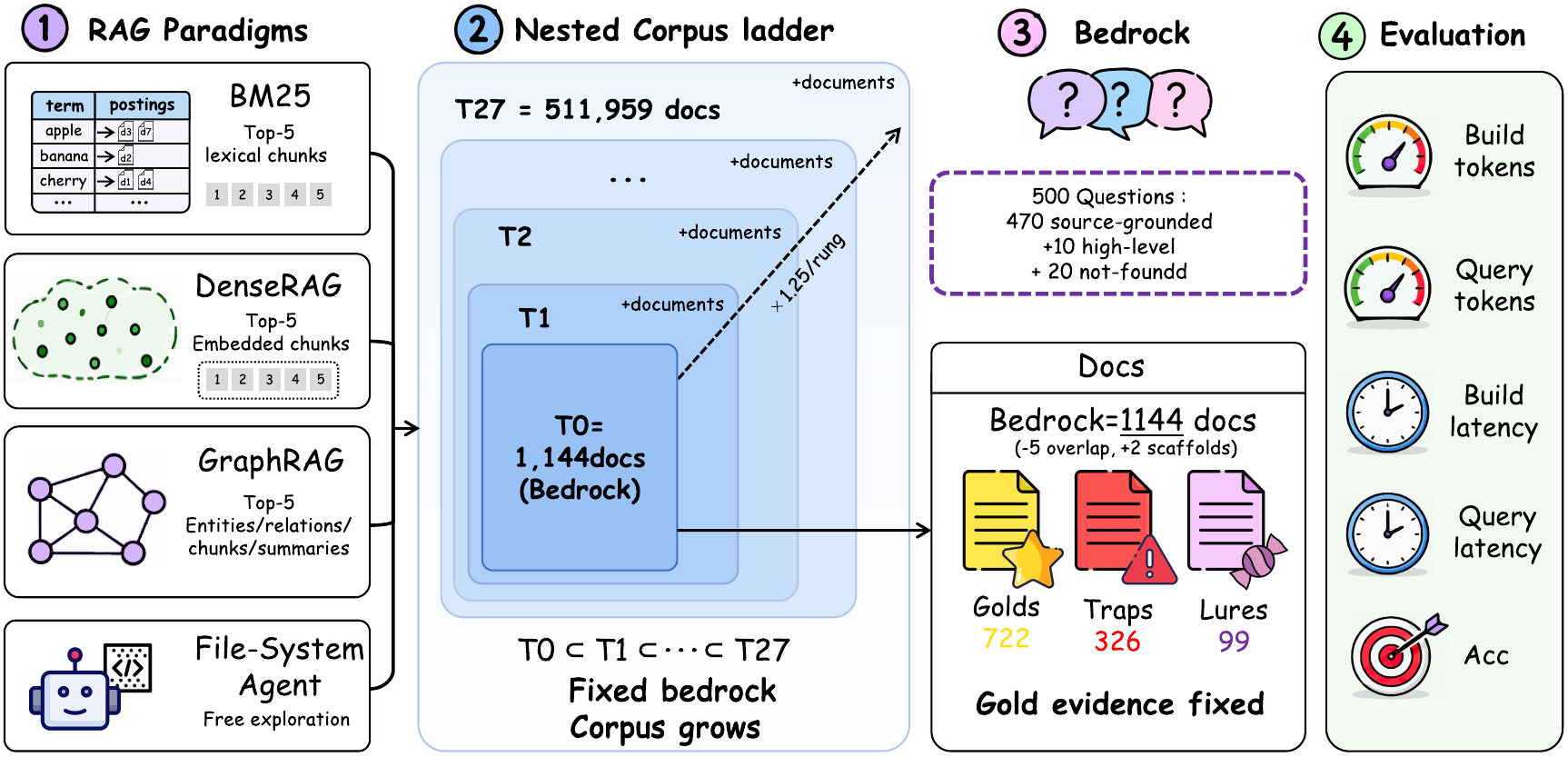}
\caption{Evaluation methodology. Four paradigms access the same nested
corpus ladder ($T_0 \subset \cdots \subset T_{27}$), which grows by
1.25 per rung while retaining a fixed 1,144-document bedrock. The 500
questions split into 470 source-grounded, 10 scaffold-supported high-level, and
20 not-found questions. Each paradigm uses the same reader and is metered
on build/query tokens, latency, and official accuracy.}
\label{fig:overview}
\end{figure}

\subsection{Benchmarks and Judging}
Existing RAG benchmarks evaluate answer
quality at a fixed corpus size and rarely meter offline
cost~\citep{yang2024cragcomprehensiverag}. We extend
EnterpriseRAG-Bench's multi-source corpus and official evaluation
\citep{sun2026enterpriseragbenchragbenchmarkcompany} with nested scaling,
unified cost accounting, and cross-paradigm comparison. We bound
judge dependence~\citep{zheng2023judgingllmasajudgemtbenchchatbot} through
dual protocols and measured cross-judge agreement.

\section{Method}
\label{sec:method}
This section describes the corpus, the nested scaling ladder, and the
unified answering and metering harness. Figure~\ref{fig:overview}
summarizes the full pipeline.

\subsection{Corpus and Questions}
EnterpriseRAG-Bench models a fictional
company that serves LLM inference. The corpus contains 511,957 documents
totaling 600.8M tokens, drawn from nine sources that include wiki pages,
chat threads, tickets, e-mail, meeting transcripts, CRM records, and code
reviews. Together with the benchmark's two organizational overview pages,
which we include as scaffolds, the full evaluation tier holds 511,959
documents. The benchmark provides 500 questions in ten types, ranging
from basic lookups to completeness, conflicting-information, and
high-level questions. Each question is annotated with gold documents, a
gold answer, and a list of atomic answer facts, with 722 gold
documents in total. The corpus natively marks 7.7\% of documents as
noise, misfiled under a wrong source or path or near-duplicated with
outdated facts, which serves as realistic filing noise. Depending on
type, questions are generated from source documents or by an agent
exploring the corpus. The released gold set was subsequently refined by
pooling candidate evidence from BM25, dense retrieval, and file-agent
search; label construction is therefore not tied to BM25 alone.

\subsection{The Bedrock}
The smallest tier is constructed deterministically
as the union of the 722
documents the benchmark annotates as gold for any of the 500 questions,
the 326 mined traps, the 99 lures, and the corpus' own two
organizational pages, a company overview and an initiative index, which
serve as scaffolds. Removing five cross-category duplicates yields 1,144.
Traps are mined through method-blind filtering: for each target question,
BM25 contributes its full-corpus top-10, while dense retrieval reranks a
BM25 top-200 pool (top-1,000 for not-found questions) and contributes ten
candidates. An LLM filter retains every
candidate that concerns the same entity or topic as a gold document while
reporting the wrong version, date, or decision. This procedure yields 326
traps under a single topicality-and-wrong-fact criterion. Lures serve the
info-not-found questions: the five most similar
candidates per question that the filter verifies cannot answer it, so
that unanswerable questions cannot be solved by the mere absence of
retrieved text. BM25 top-10 and dense reranking of a BM25 top-200 pool
contribute ten candidates each; no evaluated-system score or answer
judgment enters selection. A direct full-corpus DenseRAG audit
independently tests sensitivity to this proposal pool. Because
every adversarial document resides in the bedrock rather than being
introduced gradually, question-specific evidence and the adversarial set
remain fixed as the corpus grows through added non-bedrock documents.

\subsection{Nested Tiers}
Let $\pi$ denote a single seeded,
source-and-noise-stratified order of the non-bedrock corpus. Tier $t$
consists of the bedrock plus the first $n_t-1{,}144$ documents of $\pi$.
The source and noise distribution of each added prefix approximates the
global background corpus. Sizes follow $n_{t+1}\approx1.25\,n_t$, which yields 28
tiers. Prefix construction guarantees $T_1\subset T_2\subset\cdots$
exactly, verified through manifest checksums, and permits incremental
builders to extend an index from one tier to the next, which is also how
marginal build cost is metered. We report scale in corpus tokens rather
than document counts because document lengths differ across sources.

\section{Experiment}
\subsection{Experimental Setup}
\label{sec:protocol}

\subsubsection{Paradigms}
Our main scaling ladder evaluates seven native pipelines.
BM25 uses an inverted index
built without LLM involvement. DenseRAG performs dense retrieval over
chunk embeddings. HippoRAG~2 builds an open-vocabulary triple graph and
answers queries by linking them to facts and running Personalized PageRank.
MS-GraphRAG builds an entity graph with hierarchical community reports and
answers through its local search mode. LightRAG maintains dual-level
entity and relation indexes and answers through its hybrid mode. LinearRAG
constructs an entity co-occurrence graph using a lightweight named-entity
recognizer and the shared embedding model, without generative-LLM calls. The File-System Agent
operates without any index: an agent explores the raw per-source file tree
using read-only listing, search, and reading tools under a budget of 80
LLM calls per question. Every method that exposes a retrieval depth uses
the top-5 chunks. BM25, DenseRAG, and HippoRAG~2 operate on the same chunk
segmentation, so their retrieval quality is directly comparable.
MS-GraphRAG and LightRAG chunk internally, and their retrieved evidence is
matched back to the shared segmentation for recall computation.

\subsubsection{Reader and Metering}
All paradigms use the same
reader, Qwen3.6-27B at temperature zero, served by
vLLM~\citep{kwon2023efficient}. The File-System Agent uses the same model as its
policy, and all main-pipeline embedding calls go through a shared
Qwen3-Embedding-0.6B model.
A shared metering layer intercepts every LLM call and attributes prompt
and completion tokens to the build phase or the query phase, per paradigm
and per question. Embedding calls are accounted separately from
generative calls: because embedding requests carry no prompt template,
their token cost is counted from tokenizer inputs. DenseRAG counts are
exact; LinearRAG includes a documented 3\% estimate for its unavailable
entity-name stream. Latency is measured end to end under
single-stream conditions on an idle server.

\subsubsection{Judging}
Every prediction of every family is scored with the
benchmark's official protocol: holistic alignment of the candidate
against the gold answer, the fraction of atomic answer facts entailed by
the candidate, a combined score that gates completeness by correctness,
and document-level recall of the retrieved evidence. All main-ladder
cells were produced in one scoring session with one judge model. Each
matched control is likewise scored in a single shared session with
identical prompts and judge model; control estimates are not mixed into
the main ladder. Each system is run once per question; uncertainty
resamples questions rather than stochastic reruns. To establish that the
results do not depend on this
choice, we re-scored the predictions with an independent judge and
with a simpler binary protocol. The binary protocol preserves rankings
at all nine shared scales, while the independent judge agrees on 96.2\%
of pooled alignment verdicts; details follow below.

\subsection{Main Results}
\label{sec:results}

Table~\ref{tab:main} and Figure~\ref{fig:acc} show a scale-dependent
crossover. At the bedrock, File-System Agent and BM25 lead with 77.4 and
74.7, and their 95\% intervals (73.9--80.8 and 71.4--77.9) overlap.
File-System Agent retains the higher point estimate at the smallest measured
shared tiers; by roughly 10M corpus tokens the curves have crossed, and
BM25 leads at every larger shared tier. At full scale, BM25 retains 50.5,
versus 30.7 for File-System Agent and 29.9 for DenseRAG. Under the fixed
80-call budget used here, iterative raw-file
search resists corpus growth less well than global lexical ranking
despite many more model calls, while dense retrieval remains below both.

\begin{table}[!t]
\centering
\begingroup\small
\setlength{\tabcolsep}{2.6pt}
\begin{tabular*}{\textwidth}{@{\extracolsep{\fill}}lccccccccccc@{}}
\toprule
 & & & \multicolumn{7}{c}{Official combined score by corpus tier (\%)} & \multicolumn{2}{c}{@ bedrock (\%)} \\
\cmidrule(lr){4-10}\cmidrule(l){11-12}
Method & build & tok/q & $N{=}$1144 & 2254 & 6980 & 21614 & 42587 & 131876 & 511959 & completeness & document recall \\
\midrule
BM25 & 0 & 5.8K & 74.7 & 71.4 & \textbf{70.1} & \textbf{64.9} & \textbf{61.2} & \textbf{55.2} & \textbf{50.5} & 80.4 & \textbf{90.3} \\
File-System Agent & 0 & 226K & \textbf{77.4} & \textbf{75.4} & 69.9 & 62.6 & 58.9 & 50.9 & 30.7 & \textbf{81.6} & 86.4 \\
DenseRAG & emb.$^{\ddagger}$ & 4.9K & 58.1 & 55.7 & 51.0 & 44.2 & 40.7 & 36.0 & 29.9 & 65.4 & 73.4 \\
\midrule
\multicolumn{12}{@{}c@{}}{\colorbox{black!12}{\parbox[c][0.75em][c]{0.97\textwidth}{\centering\small\bfseries\itshape Graph-based Retrieval}}}\\
HippoRAG~2 & 7.5M & 6.5K & 66.2 & 63.1 & 58.6 & 53.8 & 50.5 & 41.0 & --- & 72.0 & 81.3 \\
LinearRAG & emb.$^{\ddagger}$ & 5.8K & 46.2 & 44.1 & 38.8 & 34.3 & 31.3 & 29.8 & --- & 52.0 & 62.6 \\
MS-GraphRAG & 35.1M & 10.4K & 45.9 & 44.0 & 38.4 & --- & --- & --- & --- & 54.1 & 57.9 \\
LightRAG & 34.6M & 8.5K & 48.0 & 42.5 & --- & --- & --- & --- & --- & 54.7 & n/a$^{\S}$ \\
\bottomrule
\end{tabular*}
\endgroup
\caption{Official main results. Build/query tokens, completeness, and
document recall are bedrock values. Combined score gates completeness by
correctness. Dashes denote unbuilt or unevaluated tiers;
Figure~\ref{fig:acc} shows every measured tier. $^{\ddagger}$ denotes
non-generative construction: embeddings for DenseRAG, and local NER plus
embeddings for LinearRAG. $^{\S}$ marks LightRAG runs that did not log
retrieved-evidence identifiers. LinearRAG tok/q averages its fixed,
stratified 100-question metered run.}
\label{tab:main}
\end{table}
\FloatBarrier

\noindent
\begin{minipage}[t]{0.455\textwidth}
\vspace{0pt}
HippoRAG~2 and LinearRAG stop at 131,876 documents, MS-GraphRAG at
8,750, and LightRAG at 2,254. Thus large-scale graph results are about
construction feasibility, not extrapolated accuracy: unsupported tiers
remain missing in the main curves and count as zero only in the
coverage-adjusted cross-scale summary.

\subsection{Build Cost and Scaling Walls}
\label{sec:build}

Figure~\ref{fig:costs} (left) and Table~\ref{tab:feas} identify
construction as the graph families' scaling wall. HippoRAG~2 is approximately linear
($b=1.01$), extrapolating to 2.9B generative tokens and roughly three
single-instance days at full scale. Yet at 155M corpus tokens its
724M-token build scores 41.0, about fifteen points below BM25.
\end{minipage}\hfill
\begin{minipage}[t]{0.515\textwidth}
\vspace{0pt}
\centering
\includegraphics[width=\linewidth]{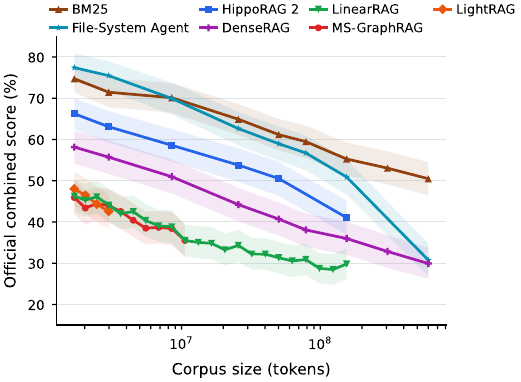}
\captionsetup{width=\linewidth,hypcap=false}
\captionof{figure}{Official combined score over the nested ladder with
95\% confidence bands. Curves cross around 10M tokens; graph curves end
at their largest feasible tier.}
\label{fig:acc}
\end{minipage}
\par\medskip

\begin{table}[H]
\centering
\begingroup\small
\setlength{\tabcolsep}{2.2pt}
\begin{tabular*}{\columnwidth}{@{\extracolsep{\fill}}lccccc@{}}
\toprule
 & $b$ & measured to & fit @full & tok/s & est. inst. \\
\midrule
HippoRAG~2 & 1.01 & 724M @ 154.7M & 2.9B & 11.9k & $\approx$3 d \\
MS-GraphRAG & 0.92 & 190M @ 10.6M & 7.9B & 1.8k & $\approx$50 d \\
LightRAG & 1.36 & 73M @ 3.0M & $\approx$102B & 0.8k & $\approx$4 y \\
LinearRAG & --- & embed only & 0 gen. & --- & $\approx$1 d \\
\bottomrule
\end{tabular*}
\endgroup
\caption{Build-cost fits $C(x)=ax^b$ against corpus tokens $x$,
extrapolated to the 600.8M-token full corpus. ``Measured to'' reports
build tokens at the largest completed corpus, and tok/s is measured
single-instance throughput; the final two columns are fitted
full-scale estimates.}
\label{tab:feas}
\end{table}

LightRAG is super-linear ($b=1.36$), does not complete at 2,826
documents within our resource envelope, and extrapolates to 102B tokens
or four instance-years. Even an optimistic $b=1.2$ leaves its estimate
near 42B tokens.

Token-based fits isolate construction work from hardware throughput.
LightRAG's repeated entity-merge rewrites yield super-linear growth; even
at three times the measured throughput, its projected full-scale build
exceeds one instance-year. Embedding-only construction is far cheaper:
DenseRAG completes the full corpus with 659.4M embedding tokens.
LinearRAG reports zero generative construction tokens, but still incurs
local NER and embedding work.

MS-GraphRAG also fits near-linear growth and completes through 8,750
documents; its full-scale estimate is 7.9B tokens and about 50
instance-days. Parallelism can reduce calendar time but not total
construction work or the resulting accuracy.

\subsection{Query Cost and Latency}

\begin{figure}[t]
\centering
\begin{minipage}[t]{0.485\textwidth}
\vspace{0pt}
\centering
\includegraphics[width=\linewidth]{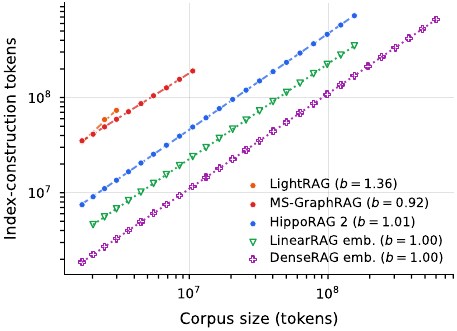}
\end{minipage}\hfill
\begin{minipage}[t]{0.485\textwidth}
\vspace{0pt}
\centering
\includegraphics[width=\linewidth]{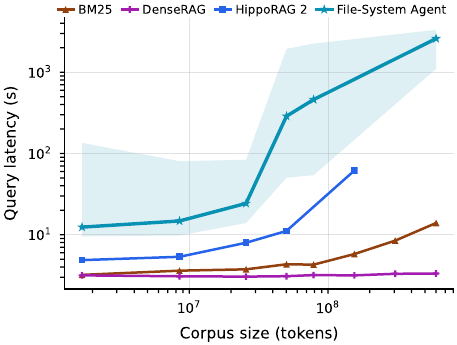}
\end{minipage}
\caption{Cost scaling across construction and querying. Left:
construction tokens and fitted power laws; hollow markers report
embedding tokens for the two non-generative builders, while BM25 and
the File-System Agent require no index. Right: single-stream query
latency; lines report medians and the File-System band spans P10--P90.}
\label{fig:costs}
\end{figure}

Figure~\ref{fig:costs} (right) reports end-to-end latency. Query cost is nearly
scale-invariant for the one-shot pipelines: BM25, DenseRAG, and
HippoRAG~2 use 5.8K, 4.9K, and 6.5K tokens per question, dominated by
the shared reader prompt. File-System Agent instead grows from 226K at
the bedrock to 343K at $N=21{,}614$, respectively 39 and 60 times
BM25 as exploration deepens.
Its median LLM calls rise from 5 to 8 by $N=42{,}587$; budget exhaustion
is below 7\% through that tier, but reaches 15\% at $N=131{,}876$ and
31\% at full scale. Accuracy also falls among within-budget questions,
so truncation alone does not explain the collapse.

Amortized over 500 questions, BM25 obtains 74.7 at 2.9M tokens, versus
77.4 at 112.8M for File-System Agent. HippoRAG~2 reaches 66.2 at 10.7M,
while MS-GraphRAG and LightRAG remain dominated near 40M tokens. The
BM25 point remains on the frontier for amortization horizons from 10
to 10,000.

These findings span the full 28-tier, roughly 450-fold ladder rather than
a single corpus size. Fixing questions, gold evidence, and distractors
isolates the effect of added background documents. Unified metering reveals
whether cost is paid offline in graph construction or online through
per-question file exploration. Across the seven native pipelines,
cross-checked judging, matched retrieval and harness controls, and the
typed graph access layer separate candidate discovery, harness, and
substrate effects. This makes the crossover a measured scaling effect
rather than a comparison of independently tuned systems at isolated sizes.

Figure~\ref{fig:summary} (left) uses six tiers and a fixed 150-question
sample. Unsupported graph tiers count as zero only for coverage
adjustment; cost averages construction and normalized 500-query
workloads over completed tiers.

\begin{table}[H]
\centering
\begingroup\small
\setlength{\tabcolsep}{3pt}
\begin{tabular*}{\columnwidth}{@{\extracolsep{\fill}}lcccc@{}}
\toprule
Tier & Files & HippoRAG~2 & LightRAG & MS-GraphRAG \\
\midrule
$N{=}1144$ & 76.181 & 75.2 (66.7) & 70.8 (48.6) & 40.9 (47.4) \\
1434 & 83.094 & 75.2 (65.9) & 75.5 (50.7) & 44.7 (44.3) \\
1798 & 76.315 & 81.2 (66.7) & 72.8 (45.5) & 46.6 (46.3) \\
2254 & 81.927 & 75.0 (65.2) & 75.6 (45.7) & 38.7 (45.4) \\
6980 & 75.111 & 70.6 (62.2) & --- & 36.9 (38.2) \\
21614 & 71.242 & 59.6 (56.0) & --- & --- \\
42587 & 64.755 & 54.2 (54.8) & --- & --- \\
\bottomrule
\end{tabular*}
\endgroup
\caption{Official score of the same agent harness over each retrieval
substrate, with native-pipeline scores on identical questions in
parentheses. Bedrock uses all 500 questions and later tiers a fixed
150-question subset for both members of each pair. Files has no native
pipeline; dashes denote unavailable matched evaluations. These scores
come from a separate matched scoring session and are not mixed with
Table~\ref{tab:main}.}
\label{tab:factorial}
\end{table}

\subsection{Behavior by Question Type}

\begin{figure}[t]
\centering
\begin{minipage}[t]{0.46\textwidth}
\vspace{0pt}
\centering
\includegraphics[width=\linewidth]{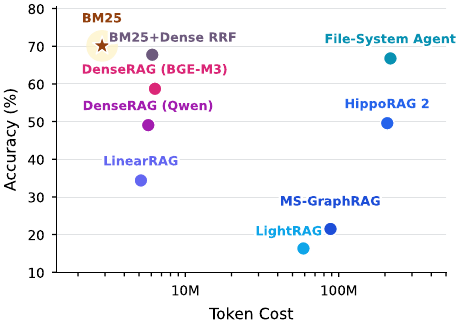}
\end{minipage}\hfill
\begin{minipage}[t]{0.52\textwidth}
\vspace{0pt}
\centering
\includegraphics[width=\linewidth]{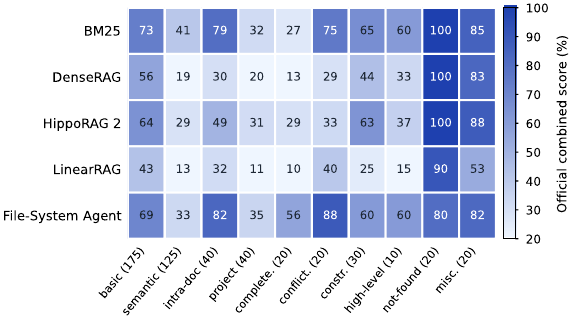}
\end{minipage}
\caption{Complementary scaling analyses. Left: cross-scale
accuracy--cost summary over six tiers on the fixed 150-question sample.
Right: official combined score by question type at $N=42{,}587$; labels
give question counts, and MS-GraphRAG and LightRAG cannot build at this
tier.}
\label{fig:summary}
\end{figure}

Figure~\ref{fig:summary} (right) shows that at $N=42{,}587$, File-System Agent
leads BM25 on intra-document, project-related, completeness, and
conflicting-information questions, including 56 versus 27 on
completeness.

BM25 is best or tied on the other five types. The saturated not-found
scores mainly measure abstention:
one-shot readers decline without supporting evidence, whereas iterative
exploration can commit to an unsupported answer. A graph system leads
only on miscellaneous.

\setlength{\emergencystretch}{1.5em}
\subsection{Cross-Protocol Robustness}
\label{sec:official}

Across nine shared scales, binary re-scoring preserves every family
ranking; official correctness is 2.60 points lower on average
(cell range $-4.60$ to $+0.69$). An independent official judge agrees on
96.2\% of pooled alignment verdicts (94.4--98.0\% per cell), with
combined-score changes from $-3.56$ to $+1.18$. Thus close pairs can vary,
but broad family separation remains. At the bedrock, BM25 retrieves any
annotated gold chunk for 94.7\% of the 470 answerable questions, all gold
chunks for 56.4\%, and averages 75.1\% gold-chunk recall at five; its
90.3\% document recall locates many residual errors within documents or in
evidence synthesis.

\FloatBarrier
\section{Separating Agency from Structure}
\label{sec:agentic}

\subsection{Retrieval Before Agency}

The full-scale BM25--File-System gap is a candidate-discovery failure,
not weak evidence synthesis. We isolate the retrieval primitive on the
same fixed, stratified 150 questions at the bedrock and full-corpus
tiers---300 question--scale pairs, not 300 unique questions. Agent+BM25
keeps the model, harness, answer constraints, question set, judge, and
80-call budget fixed; it replaces raw-tree tools with ranked lexical
search and chunk reading, together with a tool-specific instruction
adapted to that interface. Its first search is forced to use the original
question; an audit confirms that its ordered top-5 exactly matches Native
BM25 on every pair.

\begin{table}[H]
\centering
\begingroup\small
\setlength{\tabcolsep}{2.6pt}
\begin{tabular*}{\columnwidth}{@{\extracolsep{\fill}}lccccc@{}}
\toprule
& \multicolumn{2}{c}{Combined} & \multicolumn{3}{c}{At $N{=}$511,959} \\
\cmidrule(lr){2-3}\cmidrule(l){4-6}
Method & 1,144 & 511,959 & DocR & Calls & Tok./q \\
\midrule
Native BM25 & 81.3 & 54.8 & 65.6 & 1.00 & 5.8K \\
File-System & 87.1 & 36.9 & 36.8 & 36.12 & 895K \\
Agent+BM25 & 90.1 & 69.4 & 72.4 & 5.79 & 101K \\
\bottomrule
\end{tabular*}
\endgroup
\caption{Retrieval-primitive control on the same 150 questions at each
scale. Combined score and full-scale document recall (DocR) come from
one shared rejudge. Full-scale calls and tokens/question include failed
attempts that were retried. Native BM25 tokens/question is the
descriptive full-500 average.}
\label{tab:retrieval_control}
\end{table}

Table~\ref{tab:retrieval_control} shows that at the bedrock, Native BM25
and the File-System Agent are statistically tied: BM25 trails by 5.73
points, with a paired bootstrap 95\% CI of
$[-11.57,0.15]$. At the full corpus, BM25 instead leads by 17.97 points
$[9.68,26.14]$. The mechanism is discovery. Among the queries on which
each method finds at least one gold document---a descriptive,
selection-conditioned slice---the File-System Agent scores 85.9 versus
BM25's 73.8, but its any-gold hit rate collapses to 39.0\%, against
71.6\% for BM25.

Changing only the retrieval primitive reverses the collapse. At full
scale Agent+BM25 scores 69.4, beating the raw-file agent by 32.52 points
$[24.07,40.90]$ and Native BM25 by 14.56 points
$[9.22,20.12]$. Its any-gold hit rate reaches 78.0\%, while its 101K
tokens/question are roughly one ninth of raw-file exploration. Agency
therefore helps after globally ranked candidate discovery; repeated
local search is not a substitute for that global ranking. We treat this
two-scale intervention as a mechanism control, not as an eighth system
on the 28-tier native-pipeline ladder.

\begin{wraptable}{r}{0.49\textwidth}
\centering
\begingroup\small
\setlength{\tabcolsep}{4pt}
\begin{tabular*}{\linewidth}{@{\extracolsep{\fill}}lc@{}}
\toprule
Retrieval / harness & Combined \\
\midrule
File-System (ours) & 86.3 \\
File-System (Pi-Agent) & 82.3 \\
File-System (Codex) & 43.9 \\
Native BM25 & 82.1 \\
\bottomrule
\end{tabular*}
\endgroup
\captionsetup{width=\linewidth}
\caption{Same-session harness control at $N{=}$1,144 on the same 150
questions, policy model, raw corpus, and judge. Each harness retains its
native loop and stopping limit; scores are not mixed with
Table~\ref{tab:retrieval_control}.}
\label{tab:harness}
\end{wraptable}

Table~\ref{tab:harness} shows that within this matched resweep, harness
choice matters, and ours achieves the highest score of the three file
agents. Agent+BM25 therefore evaluates the retrieval swap with the
strongest observed raw-file implementation.

\subsection{Agency over Graph Substrates}

The File-System result confounds agency with its raw-file substrate. We
separate them through a paradigm-neutral access layer: each resident graph
index exposes typed, read-only tools for semantic search, neighborhood
expansion, Personalized PageRank, and chunk reading, plus its native
one-shot ranker as a tool. The same tool-calling harness is used with the
model, budget, and judge fixed; substrate-specific tools and instructions
form the intervention. Graph agents cannot access raw files, so gains
reflect policy access to index contents. A command-line interface is
included in the artifact.

Table~\ref{tab:factorial} reports the matched comparison. Agentic access
changes native scores by 22.2--29.9 points for LightRAG (4 tiers),
$-0.6$--14.5 for HippoRAG2 (7), and $-6.7$--0.4 for MS-GraphRAG (5).
Thus the same index can support different outcomes under an agent and its
native one-shot ranker.

\section{Discussion}

\subsection{The Lexical Advantage}
Enterprise questions contain precise
lexical anchors, while traps are semantically similar but factually wrong;
exact matching is therefore advantaged. Table~\ref{tab:lexical_controls}
evaluates the ordering's sensitivity to question wording and retrieval
depth.
\begin{table}[H]
\centering
\begingroup\small
\setlength{\tabcolsep}{1.6pt}
\begin{tabular*}{\columnwidth}{@{\extracolsep{\fill}}lcccccc@{}}
\toprule
Control & $N$ & BM25 & FS & Dense & Hippo2 & MS-GR \\
\midrule
\multirow{2}{*}{Paraphrase} & 1,144 & 63.9 & 73.3 & 51.8 & -- & -- \\
 & 2,254 & 60.1 & 67.2 & 49.9 & 54.9 & 36.3 \\
\midrule
\multirow{2}{*}{Top-10} & 1,144 & 83.0 & -- & 70.0 & -- & -- \\
 & 2,254 & 81.9 & -- & 66.5 & 73.0 & -- \\
\bottomrule
\end{tabular*}
\endgroup
\caption{Wording and depth controls: official combined score (\%) on
$n=148$--150; dashes denote unmeasured pairs.}
\label{tab:lexical_controls}
\end{table}
These controls leave BM25 above dense and graph retrieval under comparable
retrieval depth. In a 90-question proposal-sensitivity audit, direct
full-corpus DenseRAG top-10 retrieval overlaps the historical
BM25-prefiltered dense candidates by 1.2 documents on average, recovers
57 of 431 confirmed items, and identifies 115 traps plus 100 not-found
lures. Together, these controls show that BM25's advantage persists
under altered wording and matched retrieval depth, while adversarial
candidates are also recoverable through a direct dense proposal path.

\subsection{Graph Failure Modes}
Three factors explain the shortfall.
Construction adds extraction noise---the bedrock's 32K extracted entities
include malformed fragments---and prohibitive scale cost. Graph retrieval
also favors semantically related but factually wrong neighborhoods, which
the traps punish, while HippoRAG~2 discards extracted predicate text and
therefore relation semantics. LinearRAG, with no generative build calls,
finishes within 1.8 points of MS-GraphRAG and LightRAG at the bedrock,
suggesting that LLM extraction adds cost faster than signal here.

\subsection{Scaling Mechanism}
The decisive factor is corpus-wide candidate
discovery. BM25 and DenseRAG amortize global ranking in an index; lexical
matching also rejects this corpus' factually wrong semantic traps more
effectively. File-System Agent instead explores a local tree sequentially,
making relevant branches harder to reach as the corpus grows. Graph
methods require corpus-wide extraction, where cost, noise, and information
loss become bottlenecks. At full scale, Agent+BM25 isolates this mechanism:
changing retrieval cuts calls from 36.12 to 5.79 and tokens from 895K to
101K per question, while document recall rises from 36.8 to 72.4 and score
from 36.9 to 69.4. Iteration helps most after global ranking, not in place
of it.

\subsection{Evaluation Implications}
A single corpus size can conceal the
crossover, and graph systems can be compared only where their indexes
complete. Cross-paradigm evaluations should report nested-scale accuracy,
construction/query cost, and index coverage. Measured cells describe
quality conditional on completion; the coverage-adjusted summary in
Figure~\ref{fig:summary} (left) describes deployability, assigning zero only
to unavailable tiers. Reporting both avoids treating an unbuilt index as
an observed answer failure.

\subsection{Practical Guidance}
For enterprise-style corpora, BM25 is the
appropriate default; aggregation-heavy questions favor Agent+\allowbreak BM25, which
uses lexical ranking for discovery and agentic calls on narrowed
candidates. LLM-built graph indexes are difficult to justify at
$10^5$--$10^6$ documents unless construction is near-linear and relational
questions dominate.

\section{Conclusion}
We presented a controlled scaling study of four RAG paradigms on a
28-tier enterprise corpus ladder, holding questions, relevant evidence,
and adversarial distractors fixed while unifying the reader model, judge,
and cost metering. Matched access-layer and retrieval-swap controls further
separate substrate from agentic policy. The result is a scale-dependent
crossover: raw-file agency leads at the smallest measured shared tiers,
BM25 catches it around 10M corpus tokens, and its lead widens to nearly
20 points at 601M
tokens while remaining Pareto-optimal without LLM-based construction.
Replacing raw search with BM25 raises the full-scale agent from 36.9 to
69.4 at roughly one ninth of its query tokens, whereas graph-based RAG
encounters construction walls. Global lexical ranking therefore becomes
the stronger default as enterprise corpora scale, with agentic reasoning
best applied after candidate ranking.

\bibliographystyle{plainnat}
\bibliography{references}

\clearpage
\appendix
\section{Scope}

This supplement provides additional experimental detail, result tables,
and reproducibility notes for the main paper. The main paper is
self-contained; this document expands the protocol and gives additional
audit material. All identifiers are anonymized, and all local machine
paths are omitted.

\section{Corpus Ladder}

The study uses EnterpriseRAG-Bench, a public synthetic enterprise
benchmark with slightly more than 500K documents and 500 questions. The
scaling ladder is built by first fixing a bedrock tier that contains all
question-relevant documents, hard negatives, and not-found lures. Every
larger tier appends a seeded, source- and noise-stratified prefix of the
remaining corpus. Thus the question set, relevant evidence, and
adversarial documents are fixed while only background corpus size grows.
Manifest checks verify exact nesting. Table~\ref{tab:supp_ladder}
summarizes every tier.

\begin{table}[H]
\centering
\small
\setlength{\tabcolsep}{5pt}
\begin{tabular*}{\textwidth}{@{\extracolsep{\fill}}rrrr|rrrr@{}}
\toprule
$N$ & Corpus tok. & Chunks & Mean doc tok. &
$N$ & Corpus tok. & Chunks & Mean doc tok. \\
\midrule
1,144 & 1.7M & 2,018 & 1472.7 & 27,097 & 32.1M & 39,374 & 1182.9 \\
1,434 & 2.0M & 2,435 & 1411.1 & 33,970 & 40.1M & 49,252 & 1181.2 \\
1,798 & 2.4M & 2,955 & 1361.3 & 42,587 & 50.2M & 61,615 & 1177.7 \\
2,254 & 3.0M & 3,606 & 1324.0 & 53,389 & 62.8M & 77,142 & 1176.4 \\
2,826 & 3.6M & 4,425 & 1288.2 & 66,932 & 78.6M & 96,635 & 1174.8 \\
3,543 & 4.5M & 5,454 & 1264.3 & 83,909 & 98.6M & 121,123 & 1174.7 \\
4,441 & 5.5M & 6,737 & 1242.4 & 105,193 & 123.5M & 151,751 & 1174.4 \\
5,568 & 6.8M & 8,374 & 1227.4 & 131,876 & 154.7M & 190,096 & 1173.4 \\
6,980 & 8.5M & 10,419 & 1217.1 & 165,327 & 194.0M & 238,321 & 1173.4 \\
8,750 & 10.6M & 12,974 & 1208.6 & 207,263 & 243.2M & 298,783 & 1173.4 \\
10,970 & 13.2M & 16,169 & 1202.2 & 259,837 & 304.9M & 374,576 & 1173.4 \\
13,753 & 16.4M & 20,185 & 1195.2 & 325,746 & 382.2M & 469,677 & 1173.4 \\
17,241 & 20.5M & 25,203 & 1190.0 & 408,373 & 479.1M & 588,566 & 1173.3 \\
21,614 & 25.6M & 31,475 & 1186.6 & 511,959 & 600.8M & 737,878 & 1173.6 \\
\bottomrule
\end{tabular*}
\caption{The 28 strictly nested corpus tiers. Corpus tokens are measured
with the shared tokenizer; chunks are produced by the shared 1,200-token
chunker with 100-token overlap for systems that use the shared chunks.}
\label{tab:supp_ladder}
\end{table}

\section{Method Configuration}

All native pipelines are evaluated under one reader and judging protocol.
The shared reader is Qwen3.6-27B served with temperature zero and
thinking disabled. The shared embedding model is Qwen3-Embedding-0.6B.
BM25, DenseRAG, and HippoRAG 2 use the same shared chunks. MS-GraphRAG
and LightRAG use their native internal chunking, and retrieved evidence
is mapped back to the shared chunk identifiers when document recall is
available. Table~\ref{tab:supp_config} lists the shared settings.

\begin{table}[H]
\centering
\small
\setlength{\tabcolsep}{3pt}
\begin{tabular*}{\columnwidth}{@{\extracolsep{\fill}}ll@{}}
\toprule
Component & Setting \\
\midrule
Chunk size & 1,200 tokenizer tokens \\
Chunk overlap & 100 tokenizer tokens \\
Retriever depth & top-5 chunks where applicable \\
Reader temperature & 0.0 \\
Reader top-p & 1.0 \\
Agent budget & 80 LLM calls/question \\
Judging unit & one prediction per method/question/tier \\
Uncertainty & question bootstrap, 10,000 resamples \\
\bottomrule
\end{tabular*}
\caption{Shared settings used across native pipelines and matched controls.}
\label{tab:supp_config}
\end{table}

\subsection{Reader Prompt}

All non-agentic pipelines use the same reader prompt. The system message
is:
\begin{suppbox}{Reader system prompt}\small
You are a retrieval-based QA assistant. Answer the QUESTION using ONLY
the information in the CONTEXT. Do not use outside knowledge or invent
facts. If the CONTEXT does not contain enough information to answer, say
so explicitly. Answer in the same language as the QUESTION; be concise.
\end{suppbox}
The user message template is:
\begin{suppbox}{Reader user-message template}\small\ttfamily
CONTEXT:\\
\{context\}\\
\\
QUESTION: \{question\}\\
\\
ANSWER:
\end{suppbox}

\subsection{File-System Agent Prompt and Tools}

The File-System Agent uses the same Qwen3.6-27B model as its policy
model. The prompt gives only read-only access to the corpus tree:
\begin{suppbox}{File-System Agent system prompt}\small
You are a research assistant over an enterprise document corpus,
organized as files under source folders (slack, gmail, jira, confluence,
google\_drive, linear, github, fireflies, hubspot). Use list\_dir to
orient, grep to locate relevant documents by keyword, and read\_doc to
read them. Answer ONLY from the documents. Be concise and factual, and
cite the relative file paths you used.
\end{suppbox}
Table~\ref{tab:supp_fs_tools} lists the three exposed operations.

\begin{table}[H]
\centering
\small
\setlength{\tabcolsep}{3pt}
\begin{tabular*}{\columnwidth}{@{\extracolsep{\fill}}lll@{}}
\toprule
Tool & Arguments & Return value \\
\midrule
\texttt{list\_dir} & relative path & up to 200 child names \\
\texttt{grep} & fixed string & up to 30 matching paths \\
\texttt{read\_doc} & relative path & first 8,000 characters \\
\bottomrule
\end{tabular*}
\caption{Read-only tools exposed to the raw File-System Agent. Path
resolution rejects traversal outside the corpus root.}
\label{tab:supp_fs_tools}
\end{table}

\subsection{Agent+BM25 Prompt and Control Guarantee}

Agent+BM25 uses the same agent harness and budget, but replaces raw
tree search with a ranked BM25 search tool. Its system message is:
\begin{suppbox}{Agent+BM25 system prompt}\small
You are a research assistant over an enterprise document corpus. Use
bm25\_search with the user's original question first. Inspect the
returned top-5 chunks carefully. If they are sufficient, answer
immediately; otherwise reformulate the search query or use read\_doc on
a returned source path to inspect more context. Answer ONLY from
retrieved documents. Be concise and factual, and cite source paths or
chunk IDs. Do not use outside knowledge.
\end{suppbox}
The first \texttt{bm25\_search} call is programmatically forced to use
the original question, regardless of the tool argument. This makes the
first returned top-5 exactly the native BM25 top-5. Later calls may use
agent reformulations and are the agentic part of the intervention.

\subsection{Official Judge Prompts}

The official combined score has two LLM-judged components and one
non-LLM document-recall component. First, a holistic answer-alignment
prompt receives the query, gold answer, and candidate answer and asks
for a JSON object with a one-sentence reason and an \texttt{aligned}
field whose value is \texttt{yes} or \texttt{no}. The prompt instructs
the judge to accept stylistic differences and extra relevant context,
but to reject contradictions, wrong quantities, or missing core parts of
the query. Second, each atomic answer fact is validated with a yes/no
prompt asking whether the candidate answer contains and is consistent
with the statement. The combined score for a question is:
\[
  \mathrm{combined}(q) =
  \begin{cases}
    \mathrm{completeness}(q), & \mathrm{aligned}(q)=1,\\
    0, & \mathrm{aligned}(q)=0.
  \end{cases}
\]
Reported cell scores are the mean combined score over the evaluated
questions. Document recall is computed by exact set overlap between
retrieved document identifiers and gold document identifiers for
answerable questions.

\section{Figure and Table Metrics}

The main ladder reports native-pipeline scores only for tiers actually
built and evaluated. Dashes in the tables therefore denote missing
native results rather than extrapolated failures.

Figure~\ref{fig:teaser} is a schematic of the observed BM25--File-System
Agent crossover. Its roughly 10M-token marker summarizes the interval
between the neighboring measured tiers where their point-estimate ordering
changes; it is a rounded regime marker rather than a fitted threshold or
significance boundary. Figure~\ref{fig:acc} reports the measured curves and
confidence bands.

Figure~\ref{fig:summary} (left) summarizes six fixed tiers
($N=1{,}144,2{,}254,6{,}980,42{,}587,131{,}876,511{,}959$) on one fixed
150-question sample. Its vertical coordinate is coverage-adjusted:
unsupported graph tiers are assigned zero only for this summary. Its
horizontal coordinate is the mean over completed tiers of construction
cost plus a normalized 500-query workload. Embedding tokens are
parameter-weighted by $0.6/27$ when combined with 27B reader or generative
tokens. The plot's point ledger is included as
\texttt{figures/teaser\_crossscale\_points.csv} in the code/data supplement.

Table 1 in the main paper reports bedrock build/query tokens and
bedrock completeness/recall alongside selected ladder scores. Table 2
fits build-token scaling laws of the form $y=a x^b$ on measured build
tokens against corpus tokens and projects to the 600.8M-token full
corpus. These projections are not used as hidden accuracy estimates; the
accuracy curves show only measured cells.

\section{Token Metering}

The metering layer records prompt, completion, and total tokens for each
LLM call and assigns the call to construction or query phases. For
systems that expose OpenAI-compatible responses, usage fields are read
directly from the response. For HippoRAG 2, build tokens are recovered
from its LLM cache metadata. For MS-GraphRAG, build tokens are recovered
from its indexing logs. Embedding calls are counted from tokenizer inputs
and reported separately from generative LLM calls. LinearRAG's zero
generative-build entry therefore means zero generative LLM construction
tokens, not zero CPU work, storage, local NER, or embedding work.
Table~\ref{tab:supp_token_sources} summarizes the accounting sources.

\begin{table}[H]
\centering
\small
\setlength{\tabcolsep}{2.6pt}
\begin{tabular*}{\columnwidth}{@{\extracolsep{\fill}}lll@{}}
\toprule
Family & Build-token source & Query-token source \\
\midrule
BM25 & zero model-token build & reader usage \\
DenseRAG & embedding tokenizer inputs & reader usage \\
HippoRAG 2 & LLM cache metadata & reader usage \\
MS-GraphRAG & indexing logs & reader usage \\
LightRAG & wrapper usage logs & reader usage \\
LinearRAG & embedding backfill & reader usage \\
File-System & zero model-token build & agent call usage \\
\bottomrule
\end{tabular*}
\caption{Token-accounting sources by family. ``Zero'' denotes zero
model-token construction cost, not zero CPU, storage, or indexing work.}
\label{tab:supp_token_sources}
\end{table}

\section{Full Native-Ladder Scores}

Table~\ref{tab:supp_scores_a} and Table~\ref{tab:supp_scores_b} expand
the main result matrix beyond the subset displayed in the main paper.
Dashes denote tiers that were not built or not evaluated for that native
pipeline.

\begin{table}[H]
\centering
\small
\setlength{\tabcolsep}{2.4pt}
\begin{tabular*}{\textwidth}{@{\extracolsep{\fill}}l@{\hspace{7pt}}cccccccccccc@{}}
\toprule
Method & 1144 & 1434 & 1798 & 2254 & 2826 & 3543 & 4441 & 5568 & 6980 & 8750 & 10970 & 13753 \\
\midrule
BM25 & 74.7 & -- & -- & 71.4 & -- & -- & -- & -- & 70.1 & -- & -- & -- \\
File-System Agent & 77.4 & -- & -- & 75.4 & -- & -- & -- & -- & 69.9 & -- & -- & -- \\
DenseRAG & 58.1 & -- & -- & 55.7 & -- & -- & -- & -- & 51.0 & -- & -- & -- \\
HippoRAG 2 & 66.2 & -- & -- & 63.1 & -- & -- & -- & -- & 58.6 & -- & -- & -- \\
LinearRAG & 46.2 & 45.3 & 46.1 & 44.1 & 42.0 & 42.5 & 40.3 & 39.0 & 38.8 & 35.5 & 35.0 & 34.8 \\
MS-GraphRAG & 45.9 & 43.4 & 44.4 & 44.0 & 42.5 & 40.4 & 38.5 & 38.7 & 38.4 & 35.5 & -- & -- \\
LightRAG & 48.0 & 46.5 & 44.3 & 42.5 & -- & -- & -- & -- & -- & -- & -- & -- \\
\bottomrule
\end{tabular*}
\caption{Official combined score (\%) for smaller native-ladder tiers.}
\label{tab:supp_scores_a}
\vspace{8pt}
\begin{tabular*}{\textwidth}{@{\extracolsep{\fill}}l@{\hspace{7pt}}cccccccccccc@{}}
\toprule
Method & 17241 & 21614 & 27097 & 33970 & 42587 & 53389 & 66932 & 83909 & 105193 & 131876 & 259837 & 511959 \\
\midrule
BM25 & -- & 64.9 & -- & -- & 61.2 & -- & 59.5 & -- & -- & 55.2 & 53.0 & 50.5 \\
File-System Agent & -- & 62.6 & -- & -- & 58.9 & -- & 56.7 & -- & -- & 50.9 & -- & 30.7 \\
DenseRAG & -- & 44.2 & -- & -- & 40.7 & -- & 38.1 & -- & -- & 36.0 & 32.8 & 29.9 \\
HippoRAG 2 & -- & 53.8 & -- & -- & 50.5 & -- & -- & -- & -- & 41.0 & -- & -- \\
LinearRAG & 33.2 & 34.3 & 32.2 & 32.2 & 31.3 & 30.5 & 30.9 & 28.7 & 28.5 & 29.8 & -- & -- \\
MS-GraphRAG & -- & -- & -- & -- & -- & -- & -- & -- & -- & -- & -- & -- \\
LightRAG & -- & -- & -- & -- & -- & -- & -- & -- & -- & -- & -- & -- \\
\bottomrule
\end{tabular*}
\caption{Official combined score (\%) for larger native-ladder tiers.}
\label{tab:supp_scores_b}
\end{table}

\section{Matched Controls}

The retrieval-primitive control replaces raw file-tree exploration with
BM25 search while holding the agent model, prompt, tools, and 80-call
budget fixed. The first search is forced to use the original question,
and an audit confirms that the returned top-5 order equals the native
BM25 top-5 on each matched pair before any further agentic reasoning.
This isolates global candidate discovery from the policy loop.

The graph-substrate control exposes each graph index through typed,
read-only tools: entity/fact/relation search, neighborhood expansion,
Personalized PageRank where supported, and chunk reading. Agents over
graph substrates cannot read raw files, so gains over native one-shot
graph rankers come from policy access to the stored index rather than
from access to extra corpus text. Table~\ref{tab:supp_controls} summarizes
what each matched control fixes and changes.

\begin{table}[H]
\centering
\small
\setlength{\tabcolsep}{3pt}
\begin{tabular*}{\textwidth}{@{\extracolsep{\fill}}
>{\raggedright\arraybackslash}p{0.14\textwidth}
>{\raggedright\arraybackslash}p{0.19\textwidth}
>{\raggedright\arraybackslash}p{0.29\textwidth}
>{\raggedright\arraybackslash}p{0.29\textwidth}@{}}
\toprule
Control & Question set & What is fixed & What is changed \\
\midrule
Agent+BM25 & same 150 at bedrock/full & model, prompt, 80-call budget, judge & raw search $\rightarrow$ BM25 top-5 tool \\
Harness & same 150 at bedrock & policy model, raw files, budget, judge & agent harness implementation \\
Graph substrate agent & matched tier samples & agent model, budget, judge & stored substrate/tools \\
Top-10 & available matched cells & reader, judge, questions & retrieval depth 5 $\rightarrow$ 10 \\
Paraphrase & fixed paraphrase subset & corpus, gold, judge & question wording \\
\bottomrule
\end{tabular*}
\caption{Matched controls used to separate retrieval substrate,
agentic policy, question wording, and retrieval-depth effects.}
\label{tab:supp_controls}
\vspace{8pt}
\begin{minipage}{0.48\textwidth}
\centering
\small
\setlength{\tabcolsep}{2.2pt}
\begin{tabular*}{\linewidth}{@{\extracolsep{\fill}}lccccc@{}}
\toprule
Control & $N$ & BM25 & FS & Dense & Graph \\
\midrule
Paraphrase & 1,144 & 63.9 & 73.3 & 51.8 & -- \\
Paraphrase & 2,254 & 60.1 & 67.2 & 49.9 & 54.9 / 36.3 \\
Top-10 & 1,144 & 83.0 & -- & 70.0 & -- \\
Top-10 & 2,254 & 81.9 & -- & 66.5 & 73.0 \\
\bottomrule
\end{tabular*}
\caption{Wording and retrieval-depth controls. Graph entries are
HippoRAG 2 and, where available, MS-GraphRAG.}
\label{tab:supp_lexical_controls}
\end{minipage}
\end{table}

\section{Robustness Checks}

The official judge uses correctness-gated completeness as the combined
score. A simpler binary protocol preserves all broad family rankings at
the nine shared native-pipeline scales. An independent official judge
agrees with the primary judge on 96.2\% of pooled alignment decisions,
with per-cell agreement ranging from 94.4\% to 98.0\%. Close pairs can
move by a few points, but the main separation between BM25, dense
retrieval, raw-file agency at scale, and graph construction families is
stable.

The lexical-overlap controls use paraphrased questions and top-10
retrieval budgets at the two smallest tiers where enough methods are
available. BM25 remains above dense and graph retrieval under comparable
retrieval depth. A proposal-sensitivity audit also performs direct
full-corpus DenseRAG top-10 retrieval for 90 questions, confirming that
trap proposals are not exclusively reachable through the BM25-prefiltered
dense pool used during benchmark construction. Exact matched-cell values
are reported in Table~\ref{tab:supp_lexical_controls}.

\section{Failure and Stopping Criteria}

For LLM-built graph methods, a tier is counted as completed only when
the construction artifact is usable for all scheduled questions and its
token ledger is available. If a build exceeds the resource envelope,
fails during entity/relation extraction or merge, or produces an
incomplete index that cannot answer the scheduled cell, the tier is
reported as unavailable rather than assigned an inferred accuracy in the
main ladder. The cross-scale summary is the only place where unavailable
graph tiers are converted to zero, and its caption labels the result as
coverage-adjusted.

File-System Agent failures are handled at question level. Failed or
retried attempts are included in the full-scale call and token totals.
Budget exhaustion is recorded separately from answer correctness. The
main text reports that accuracy also falls among within-budget questions,
so the full-scale loss is not only a truncation artifact.

\section{Data Schemas}

Native prediction files use one JSON object per question with fields:
\begin{suppbox}{Native prediction record}\small\ttfamily
id, question, predicted\_answer, retrieved\_chunk\_ids,\\
retrieved\_doc\_ids, latency\_sec, metadata
\end{suppbox}
Official judgment files bind a judgment to the answer and retrieved
evidence using a SHA256 hash and record:
\begin{suppbox}{Official judgment record}\small\ttfamily
id, aligned, reason, completeness\_pct, n\_facts,\\
fact\_results, doc\_recall, prediction\_sha256
\end{suppbox}
Token ledgers are phase separated:
\begin{suppbox}{Phase-separated token ledger}\small\ttfamily
baseline, dataset, source, build\{prompt,completion,total,calls\},\\
qa\{prompt,completion,total,calls\}, build\_qa\_total\_tokens
\end{suppbox}

\section{Bootstrap Intervals}

Confidence intervals in the native ladder resample questions with
replacement within each method--tier cell. Each interval uses 10,000
bootstrap resamples and reports the 2.5th and 97.5th percentiles.
Matched controls use paired resampling over the same question IDs so
that the confidence interval is over the paired score difference.

\FloatBarrier
\section{Artifact-to-Claim Map}

The major empirical claims map to the following code/data artifacts.
\begin{itemize}
\item Native-ladder accuracy and confidence bands:
\path{results/judge_official_canonical_20260724/}.
\item Build scaling: \path{results/*scaling_ledger.csv} and
\path{results/build_cost_authoritative.csv}.
\item Query tokens: \path{results/token_usage/}.
\item Agent+BM25: \path{results/judge_agent_bm25_control/} and
\path{results/agent_bm25_control_report.md}.
\item Graph-substrate agents:
\path{results/judge_official_table3_20260724/}.
\item Lexical controls: \path{results/judge_paraphrase_*} and
\path{results/judge_topk10_*}.
\item Cross-scale summary: \path{figures/teaser_crossscale_points.csv}.
\end{itemize}

\section{Artifact Contents}

The code/data supplement is organized as follows.

\begin{itemize}
\item \texttt{scripts/}: sanitized scripts for constructing tiers,
running native pipelines, metering tokens, judging outputs, and producing
paper figures.
\item \texttt{results/}: aggregate CSV/JSON ledgers used for the paper
tables, figures, and mechanism controls.
\item \texttt{figures/}: figure-generation inputs and final PDFs;
\texttt{README.md}: environment variables and reproduction order.
\end{itemize}

\end{document}